\documentclass[runningheads]{llncs}
\usepackage{graphicx}
\usepackage{amsmath,amssymb} 
\usepackage{color}

\newcommand{\hstate}{\mathbf{h}}

\usepackage[width=122mm,left=12mm,paperwidth=146mm,height=193mm,top=12mm,paperheight=217mm]{geometry}
\begin{document}
\pagestyle{headings}
\mainmatter

\title{Learning Language-Visual Embedding for Movie Understanding with Natural-Language} 

\titlerunning{A very long title}

\author{Atousa Torabi \inst{1} \and Niket Tandon \inst{2} \and Leonid Sigal \inst{1} }

\institute{Disney Research Pittsburgh \and
                      Max Planck Institute for Informatics\\
                      \email{ \{atousa.torabi,lsigal\}@disneyresearch.com}\\
                      \email{ntandon@mpi-inf.mpg.de}}


\maketitle

\begin{abstract}
Learning a joint language-visual embedding has a number of very appealing properties and can result in variety of practical application, including natural language image/video annotation and search. In this work, we study three different joint language-visual neural network model architectures. We evaluate our models on large scale LSMDC16 \cite{LSMDC15web,LSMDC15} movie dataset for two tasks: 1) Standard Ranking for video annotation and retrieval 2) Our proposed movie multiple-choice test. This test facilitate automatic evaluation of visual-language models for natural language video annotation based on human activities. In addition to original Audio Description (AD) captions, provided as part of LSMDC16, we collected and will make available a) manually generated re-phrasings of those captions obtained using Amazon MTurk  b) automatically generated human activity elements in "Predicate + Object" (PO) phrases based on "Knowlywood", an activity knowledge mining model \cite{Tandon2015Knowlywood}. Our best model archives Recall@10 of 19.2\% on annotation and 18.9\% on video retrieval tasks for subset of 1000 samples. For multiple-choice test, our best model achieve accuracy 58.11\% over whole LSMDC16 public test-set.  

\end{abstract}

\section{Introduction}
Natural language-based video and image search has been a long standing topic of research among information retrieval, multimedia, and computer vision communities. Several existing on-line platforms (e.g. Youtube) rely on massive human curation efforts, manually assigned tags, click counts and surrounding text to match largely unstructured search phrases in order to retrieve ranked list of relevant videos from a stored library. However, as the amount of unlabeled video content grows, with advent of inexpensive mobile recording devices (e.g. smart phones), the focus is rapidly shifting to automated understand, tagging and search.   

Over the last 1-2 years, there has been an increased interest in jointly modeling images/videos and natural language sentences. Models that jointly learn from videos/images and natural language sentences have broad applicability to visual search, retrieval \cite{kiros15tacl,IvanVendrov2015}, captioning \cite{donahue15cvpr,XuBKCCSZB15,vinyals15cvpr}, or visual question answering \cite{VQA,gao2015mQA,948} tasks. There are also abundant evidence \cite{Sadeghi2015,Wu2016} that jointly learning from language and visual modalities can be mutually beneficial. These recent trends of multi-modal learning are, at least in part, are enabled by recent advances in deep learning, with high capacity Convolutional Neural Networks (CNNs) driving up performance on image and video recognition \cite{SimonyanZ14,Simonyan14c,WuWJYX15} end, and Recurrent Neural networks (RNNs), particularly Long Short-Term Memory (LSTM), making advances in natural language translation and sentence generation. The learning of such high capacity models, with millions and tens of millions of parameters, is made possible by availability of large scale image- (e.g. COCO \cite{502}, Visual Genome\footnote{\url{https://visualgenome.org/}}) and, more recently, video- (LSMDC \cite{LSMDC15}, MSR-VTT \cite{264836}) description datasets, where an image or video is associated with one-or-more human generated natural language sentence descriptions.

While there are a number of recent works that look at joint image-language modeling, the progress on {\em video}-language models has been much more limited. The difficulty stems from additional challenges that come from the need to encode the temporal aspects of the video and the sheer volume of the video data required to be processed (e.g. Large Scale Movie Description Challenge (LSMDC) \cite{LSMDC15} dataset contains nearly 60-times the number of frames as images in COCO \cite{502}). In the video-language domain, the few works that exist focus largely on video description generation \cite{VenugopalanRDMD15,yao2015capgenvid}. These models typically use LSTM models to generate sentences given an encoding of the video. Evaluating the video description performance  especially on LSMDC that contain audio description (AD) captions and are typically relatively verbose and very precise is not easy and usually have done based on human judgment. In this work we study two tasks: 1) Standard Ranking for video annotation and retrieval 2) Multiple-choice test, which enable us to automatically evaluate joint language-visual model based on precise matrics: recall and accuracy. We provide a baseline results for LSMDC16 video annotation and search and multiple-choice test data, that we built for LSMDC16.

The rest of the paper is organized as the following, in section \ref{relatedwork}, we describe exisiting joint visual-language learning related works, in section \ref{multiplechoice}, we explain how we built LSMDC16 multiple-choice test. Section \ref{2stream} describes our joint visual-language models structure and our experiments is presented in section \ref{experiment}. 

\section{Related Work}
\label{relatedwork}
\noindent
{\bf Visual Caption Generation:}
There is an increasing interest in jointly learning from images/videos and natural language sentences. In the last 1-2 years, several works have been developed for automatic image \cite{chen15cvpr,donahue15cvpr,fang15cvpr,karpathy15cvpr,kiros15tacl,XuBKCCSZB15,vinyals15cvpr} and video \cite{VenugopalanRDMD15,yao2015capgenvid} captioning. Most of these models use Recurrent Neural Networks (RNN), or LSTMs, in an encoder-decoder architecture (i.e. CNNs or CNN+LSTM encoder, for images or videos respectively, to generate a hidden semantic state and RNN/LSTM decoder that decodes resulting hidden state using a trained language model to produce a final sentence description). Automatic captioning is a very challenging task, both from an algorithmic and evaluation point of view. The latter is particularly challenging with difficulties arising from evaluation of  specificity, linguistic correctness and relevance of generated captions. 


\noindent
{\bf Visual Question Answering:}
Because of the aforementioned challenges, image \cite{VQA,gao2015mQA,948} and, more recently, video \cite{Linchao2015} Visual Question Answering (VQA) has became a preferred alternative to caption generation. VQA datasets \cite{VQA,gao2015mQA,tapaswi16cvpr,Linchao2015} consist of large number of structured (e.g. multiple choice or fill-in-the-blank) and unstructured (e.g. free form) image-specific questions with corresponding answers. As a result, evaluation in VQA setting tends to be considerably more objective, requiring algorithms to have, at a minimum, certain level of visual understanding to answer the poised questions. 

\noindent
{\bf Image-Caption Retrieval:}
Another alternative is image-caption retrieval that has been defined as a standard way to evaluate joint language-visual models \cite{hodosh13jair,kiros15tacl,lin14cvpr,IvanVendrov2015}. The core idea is to rank a set of images according to their relevance to a caption query ({\em a.k.a}, image retrieval) or ranking captions according to their relevance to the given image query ({\em a.k.a}, caption retrieval). 
Image-caption retrieval approaches, typically, learn a joint embedding space that minimizes a pair-wise ranking objective function between images and captions. Particularly relevant to our paper, is the work of \cite{IvanVendrov2015}, where it is acknowledged that different forms of captions form a visual-semantic hierarchy and the order-preserving constraints are used as objective for learning.
\begin{figure}[ht]
\begin{center}
   \includegraphics[width=0.8\linewidth]{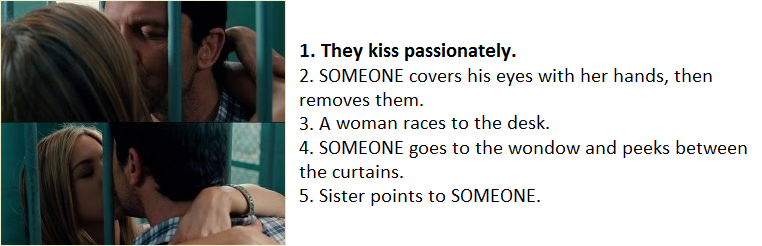}
\end{center}
\vspace{-0.2in}
   \caption{ An example of LSMDC16 multiple-choice test.}
\label{overview}
\label{fig:Multiplechoice}\
\vspace{-0.5in}
\end{figure}

\section{LSMDC16 Multiple-Choice Test}
\label{multiplechoice}
In this section we explain our multiple-choice test collection for LSMDC16.\\

\noindent
{\bf LSMDC16 multiple-choice test definition:} Given a video with 5 possible description choices, the task is assigning correct(ground-truth) caption to the video among 5 possible choices. Figure \ref{fig:Multiplechoice} illustrate multiple-choice test for a video sample. 

\noindent
In order to define 5 possible captions for each video, first we tag each caption in the whole LSMDC16 with one or multiple activity phrase labels (i.e. described in the following paragraph). Second the correct answer is the ground-truth caption and four other disctractor answers are randomly picked from the same subset (i.e. either training, test, or validation captions) with the condition that their activity phrase labels have no word intersection with correct answer activity phrase labels.

\begin{figure}[ht]
\begin{center}
   \includegraphics[trim = 0mm 2mm 0mm 0mm, clip, width=0.8\textwidth]{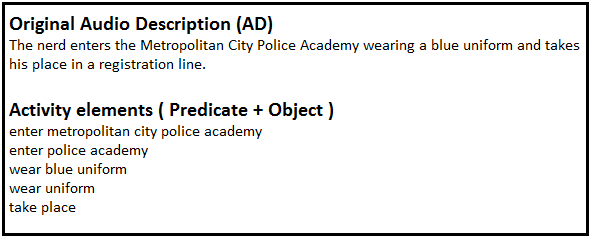}
\end{center}
\vspace{-0.2in}
   \caption{ An Example of Activity phrase mining from movie AD.}
\label{overview}
\label{fig:Activityphrase}\
\vspace{-0.2in}
\end{figure}

\noindent
{\bf Activity knowledge mining:} The activity elements
are extracted from AD captions using "Knowlywood", a model for human activities mining from movie narratives such as AD captions \cite{Tandon2015Knowlywood} and it is presented in "Predicate + Object" (PO) format. Figure \ref{fig:Activityphrase} illustrates an example of activity phrase labels from a AD caption (for more details about activity mining please see \cite{Tandon2015Knowlywood}).

\noindent
{\bf Multiple-choice test evaluation:} For evaluation, we compute test accuracy for public-test captions with 10053 videos. Test accuracy is the percentage of correctly answered questions (i.e. 10053 questions).


\begin{figure}[]
\begin{center}
    \includegraphics[trim = 0mm 15mm 0mm 0mm, clip, width=1.02\textwidth]{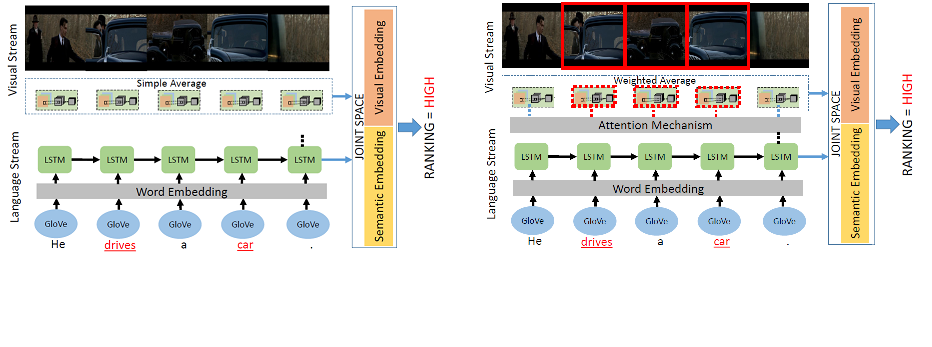} 
    \textbf{M2}  ~~~~~~~~~~~~~~~~~  ~~~~~~~~~~~~~~~~~ ~~~~~~~~~~~~~~~~~ \textbf{M3}  ~~~~~~~~~~~~~
\end{center}
\vspace{-0.1in}
   \caption{The variations of neural networks models for learning joint visual-language embedding.}
\vspace{-0.5in}
\label{Arc}   
\end{figure}

\section{ Joint visual-language neural network models}
\label{2stream}

Our overall goal is to learn, given a set of training video-caption pairs $\mathcal{D}_{vid} = \{ ( \textbf{V}_i, \textbf{S}_i ) \}$, a ranking function such that the ground-truth pairs $(\textbf{V}_i, \textbf{S}_i)$ rank higher than all other pairs (e.g. $(\textbf{V}_i, \textbf{S}_j)$ where $j \neq i$) by a margin. While this high level goal is easy to specify it requires a number of algorithmic choices and challenges to realize. we implement three different models: 

\noindent
\textbf{M1. Simple Average Glove + Simple Average FC-7 (SA-G + SA-FC7)}:  We use this simple model as the baseline. In this model the sentence is encoded using simple average word vector representations. Also video is encoded using simple average of video frames feature vectors as we will describe in Appendix~\ref{input}. Finally the video representation and sentence representation are separately linearly transformed into joint visual-language space. We constrain video and sentence embedding vectors to only positive values by computing their absolute values as suggested in \cite{IvanVendrov2015}. The model is trained using a ranking objective function that will be described in Appendix~\ref{LR}.

\noindent
\textbf{M2. Language LSTM + Simple Average FC-7 (LSTM + SA-FC7)}:
Figure \ref{Arc} (left) illustrates this model. In this model the sentence is encoded using LSTM \cite{Graves13} and represented in joint semantic space as LSTM last hidden layer output; video is encoded using simple average of video frames feature vectors as we will describe in Appendix~\ref{input}. Finally the video representation is linearly transformed into joint semantic space. In this model, we also constrain video and sentence embedding vectors to only positive values by computing their absolute values. The model is trained using a ranking objective function that will be described in Appendix~\ref{LR}.

\noindent
\textbf{M3. Language LSTM + Weighted Average FC-7 (LSTM + WA-FC7)}: Figure \ref{Arc} (right) illustrates this model. In this model, similar to above, the sentence is encoded using LSTM and its embedding is absolute values of last hidden state output of LSTM. A soft-attention aligns the output of last hidden state of language LSTM, $\hstate_{N}$, with all feature vectors of frames in $\mathbf{V}$. Video is then encoded using weighted average of frame feature vectors and the video representation is linearly transformed into joint embedding space obtained by minimizing a ranking loss that we will describe in Appnedix \ref{LR}.

\section{Experiments}
\label{experiment}
\noindent
{\bf Datasets:} We use two datasets in our experiments:  
\begin{enumerate}
\item COCO image description dataset \cite{502} with 123,287 images: 113,287 training images, 5,000 test and 5,000 validation images. Each image has 5 ground-truth captions. 
\item LSMDC16 \cite{LSMDC15} movie description dataset with 101,079 training, 10,053 test and 7,408 validation videos. Each video has only one ground-truth caption.
\item LSMDC16 \cite{LSMDC15} rephrases, we have collected 17,000 captions generated by human using Amazon's Mechanical Turk (AMT) for LSMDC16 datasets. 11,000 captions are from training set and 6,000 captions are from test set. We asked people to re-phrase original captions with 3-10 word sentences using different wording compared to original captions as much as possible. As an example, ``She walks over to the banquette and sits down." is original caption was re-phrased as ``She walked to the feast and took a seat.".
\end{enumerate}
\subsection{Implementation details}
 We implemented our models in Theano-based framework named BLOCKS \cite{Bastien-Theano-2012,MerrienboerBDSW15}. Our data pipeline has been implemented using FUEL framework \cite{MerrienboerBDSW15}.
 
\noindent
{\bf Model configurations:} We set LSTMs hidden-state dimension and the dimensions of the joint embedding space to $950$ (we also tried $512$ but it resulted in lower performance). The dimension of the word embedding is $300$. The attention model matching space is also $300$ (we also tried $900$ which resulted in similar performance), and set the ranking margin $\alpha = 0.05$ (as proposed in \cite{IvanVendrov2015}). Similar to other works, e.g. \cite{kiros15tacl,IvanVendrov2015}, to decrease overfitting issue, we constrain sentence and video embeddings to have a unit (L2) norm. 

\noindent
{\bf Training:} 
We found it useful to train with both video-sentence (LSMDC) and image-sentence (COCO) datasets; to accommodate this for standard ranking task we treat images as one-frame videos. For both standard ranking and multi-choice test training phase is the same. However for multi-choice test we tried both pairwise ranking and annotation ranking as we described in Appendix \ref{LR}. In training phase, we randomly sample minibatch of size $200$ image/video-caption pairs. The contrastive terms for the minibatch are computed by using $199$ contrastive videos for each caption and $199$ contrastive captions for each video, similar to \cite{IvanVendrov2015}. Our models were trained with Adam optimizer with learning rate $0.001$ and we applied clipping gradient with threshold $2.0$. We early-stop and save the best model based on monitored loss function of $1000$ randomly selected validation samples. For all datasets, including LSMDC16 and COCO, we use original train, valid, and test splits.   

{\bf Testing:} For evaluation standard video annotation and retrieval, we use Recall@K and Median Rank (medR) metrics. ``r@k" means the percentage of ground-truth captions/videos in the first \textit{K} retrieved captions/videos and ``medR" means the median ranks of ground-truth videos/captions. We report results on COCO test images, LSMDC16  test set, and LSMDC16 test rephrasing captions that we will make public. For multi-choice test evaluation, we compute the accuracy over test-data. In order to do that for each test query, first we compute order similarity (Equation \ref{eq:sim} in Appendix~\ref{LR}) for all 5 answer choices and then video is assigned to description with highest score, then accuracy is computed for all test data as we described in Section \ref{multiplechoice}.

\subsection{Qualitative Results for temporal attention network}
\label{qualitative}

\begin{figure}[ht]
\begin{center}
\includegraphics[width=0.9\textwidth]{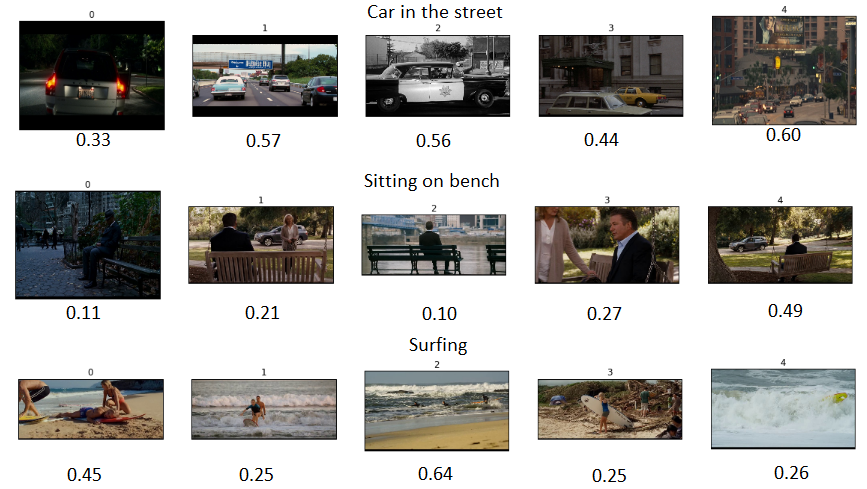} \\
\end{center}
\vspace{-0.2in}
 \caption{Top 5 (from left to right) phrase-based video search results. Each row shows video frmaes with maximum attention weight computed with model \textbf{M3} (\textit{C+L'16}). The phrase query is on top of each row. The value below each image is the attention weight for corresponding query.}
\label{overview}
\label{fig:equals}\
\vspace{-0.1in}
\end{figure}
\noindent
Figure \ref{fig:equals} shows results of top 5 ( at each row, from left to right) phrase-based video search. A phrase query have been shown on top of each row. Each row shows the video frame with highest attention weight from the retrieved video computed for corresponding phrase query. In order to compute attention weights for a new phrase query, we forward pass top 5 videos, retrieved by model \textbf{M1} or \textbf{M2}, and new phrase query through attention network in \textbf{M3} model, trained on combination of COCO and LSMDC. We observe that our temporal attention network is able to highlight salient video frames in LSMDC video clips by associating a higher attention weights to salient video frames.

\subsection{Quantitative Results}
\label{quantitative}

\begin{table*}[!t]
\begin{center}
\caption{Quantitative comparison of models for standard ranking. \textbf{M1} is \textit{SA-G + SA-FC7} model and \textbf{M2} is \textit{LSTM + SA-FC7} (details are in Section~\ref{2stream}), and \textbf{OE} stands for order-embeddings model \cite{IvanVendrov2015}. The \textbf{C} stands for \textit{COCO} image dataset, \textbf{L'16} for \textit{LSMDC'16} datasets, and  \textbf{RP} for \textit{LSMDC'16} rephrases data. The \textbf{1000 AD} stands for 1000 randomly picked audio description captions from LSMDC16 public test-set; \textbf{1000 re-phrase} stands for 1000 captions re-phrasings from LSMDC16 public test-set.}
{ \small
\begin{tabular}{|l|c|c|c|c|c|c|c|c|c|}
\hline
 \multicolumn{1}{|c|} {Model} & \multicolumn{4}{|c|} {Annotation (1000 coco test)} & \multicolumn{4}{|c|} {Retrieval (1000 coco test)}\\
 &R@1 & R@5 & R@10 & medR &R@1 & R@5 & R@10 & medR\\
 \hline
$\textbf{M2}_{}$ (C) & 44.6 & 78.4 & \textbf{89.5} & 2 & 37.2 & 72.3 & \textbf{85.9} & 2 \\
\hline
 $OE_{}$(C)\cite{IvanVendrov2015} & \textbf{46.7} & - & 88.9 & 2 & \textbf{37.9} & - & \textbf{85.9} & 2\\
\hline
\multicolumn{9}{c}{} \\
\hline
 \multicolumn{1}{|c|} {Model} & \multicolumn{4}{|c|} {Annotation (1000 AD )} & \multicolumn{4}{|c|} {Retrieval (1000 AD )}\\
 \hline
 $\textbf{M1}_{}$ (C+L'16) & 2.4 & 9.0 & 14.0 & 113 & 3.0 & 8.8 & 13.2  & 114 \\
 \hline
 $\textbf{M2}_{}$ (L'16) & 3.5 & 10.8 & 17.1 & 98 & 3.3 & 10.2 & 15.6  & \textbf{88} \\
\hline
 $\textbf{M2}_{}$ (C+L'16) & \textbf{4.0} & 12.2 & \textbf{19.2} & 91 & \textbf{4.3} & 12.6 & 18.9  & 98 \\
\hline
$\textbf{M2}_{} $(C+L'16 + RP) & 3.9 & \textbf{12.6} & 18.1 & \textbf{90} & 4.2 & \textbf{13.0} & \textbf{19.5}  & 90 \\
\hline

\multicolumn{9}{c}{} \\
\hline
\multicolumn{1}{|c|} {Model} & \multicolumn{4}{|c|} {Annotation (1000 re-phrase) } & \multicolumn{4}{|c|} {Retrieval (1000 re-phrase)}\\
\hline
 $\textbf{M1}_{}$ (C+L'16) & 1.8 & 6.7 & 11.0 & 149 & 2.0 & 6.8 & 10.8  & 155 \\
 \hline
 $\textbf{M2}_{} $(L'16) & 2.4 & 6.9 & 11.8 & 139 & 2.0 & 6.9 & 11.8  & 131 \\
\hline
 $\textbf{M2}_{}$ (C+L'16) & \textbf{2.6} & \textbf{8.5} & \textbf{12.2} & 137 & \textbf{2.5} & \textbf{7.9} & \textbf{12.7}  & 138 \\
\hline
 $\textbf{M2}_{} $(C+L'16+RP) & 2.2 & 7.0 & 12.1 & \textbf{124} & 1.8 & \textbf{8.1} & 12.5  & \textbf{124} \\
\hline
\end{tabular}
}
\label{tab1}
\end{center}
\vspace{-0.2in}
\end{table*}

\noindent
{\bf Standard ranking:}
Table \ref{tab1} summarizes all our evaluations for  \textbf{M1} and \textbf{M2} model architectures described in Section~\ref{2stream}. The best results are shown in bold. 
Following are some observations. First, using model \textbf{M2}, we are able to reproduced COCO image retrieval SOTA results in \cite{IvanVendrov2015} (see row 1 and 2). Second, \textbf{M2} (\textit{C+L'16}) trained on combination of COCO (C) and LSMDC16 (L'16), has consistently a better performance, for both 1000 original captions and re-phrases, compared to the same model only trained on LSMDC16 (L'16) or baseline model \textbf{M1} (\textit{C+L'16}). Finally model \textbf{M2} (\textit{C+L'16+RP}) trained on combination of COCO (C), LSMDC16 (L'16), and LSDMC16 re-phrases has better medR performance compared to \textbf{M2} (\textit{C+L'16}).   

Even the Standard ranking performance for original AD and rephrased captions is relatively low due to the caption complexity, still with simple phrase queries, we can obtain quiet impressive video retrieval results using trained joint language-visual models for LSMDC16 (see more qualitative results in Appendix \ref{App:AppendixA}).

\begin{table*}[!t]
\begin{center}
\caption{Quantitative comparison of models for multiple-choice test. \textbf{M1} is \textit{SA-G + SA-FC7} model and \textbf{M2} is \textit{LSTM + SA-FC7} (details are in Section~\ref{2stream}). The \textbf{{AR}} stands for annotation ranking (details in Appendix~\ref{LR}). The \textbf{C} stands for \textit{COCO} image dataset, \textbf{L'16} for \textit{LSMDC'16} datasets, and  \textbf{RP} for \textit{LSMDC'16} rephrases data.}
{ \small
\begin{tabular}{|l|c|c|c|c|c|c|c|c|c|}
\hline
 \multicolumn{1}{|c|} {Model} & \multicolumn{1}{|c|} {multiple-choice (10053 LSMDC16 test)} \\
 &accuracy\\
\hline
$\textbf{M1}_{}$ (C+L'16) & 55.1\% \\
\hline
$\textbf{M2}_{}$ (L'16) &  56.3\%\\
\hline
$\textbf{M2}_{}$ (C+L'16)  & 56.6\%\\
\hline
$\textbf{M2}_{}$ (C+L'16+AR) & \textbf{58.1}\%\\
\hline

\end{tabular}
}
\label{tab2}
\end{center}
\vspace{-0.2in}
\end{table*}

\noindent
{\bf Multiple-choice test:}
Table \ref{tab2} summarizes all our evaluations for  \textbf{M1} and \textbf{M2} model architectures described in Section~\ref{2stream}. The best result are shown in bold.
First, \textbf{M2} (\textit{C+L'16+AR}) which is trained on combination of COCO and LSMDC has the best performance. Second, training a model using \textbf{AR} (Annotation Ranking) loss has $2\%$ improvement in accuracy compared to pairwise ranking for multiple-choice test (see ranking loss details in Appendix \ref{LR}). 

\clearpage
\bibliographystyle{splncs03}
\bibliography{egbib}
\newpage
\appendix

\section{Phrase-based movie shots search}\label{App:AppendixA}

\begin{figure*}[!hb]
\begin{center}
\includegraphics[width=0.97\textwidth]{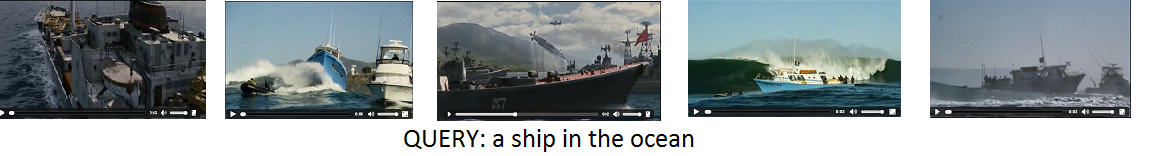} \\
\includegraphics[width=0.99\textwidth]{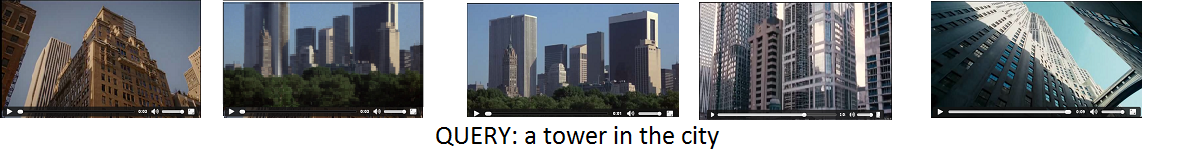} \\
\includegraphics[width=0.97\textwidth]{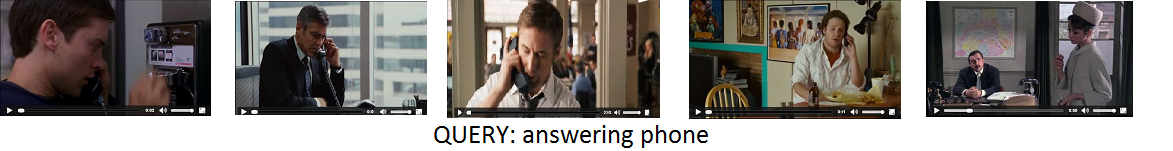} \\
\includegraphics[width=0.97\textwidth]{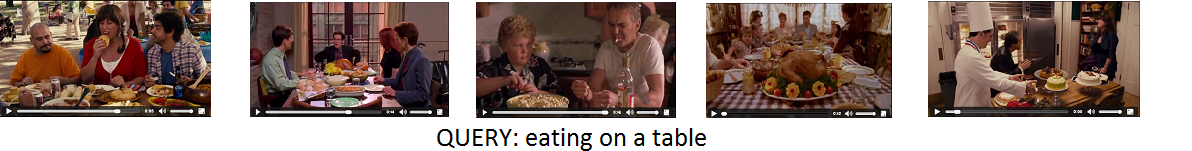} \\
\includegraphics[width=0.99\textwidth]{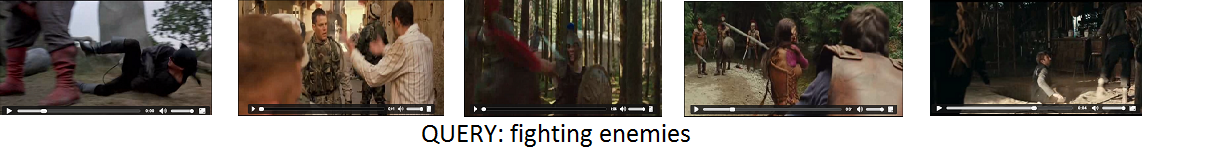} \\
\includegraphics[width=0.95\textwidth]{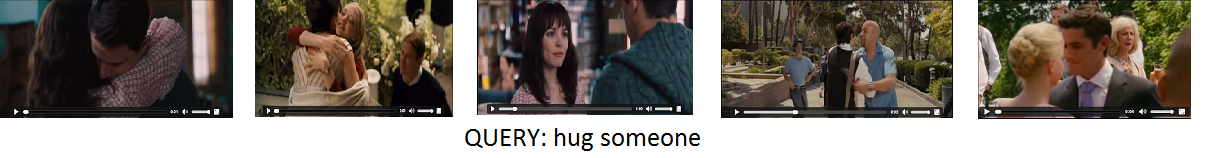} \\
\includegraphics[width=0.95\textwidth]{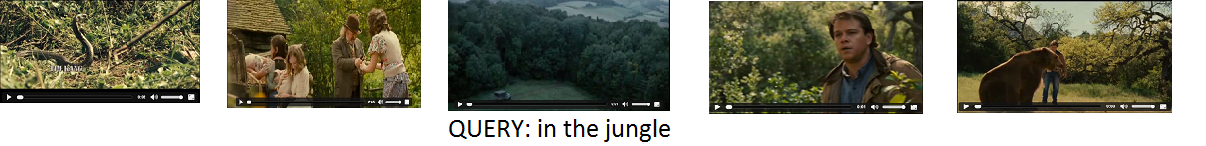} \\
\includegraphics[width=0.94\textwidth]{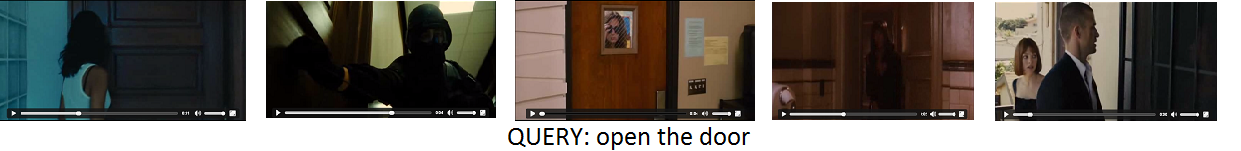} \\
\includegraphics[width=0.99\textwidth]{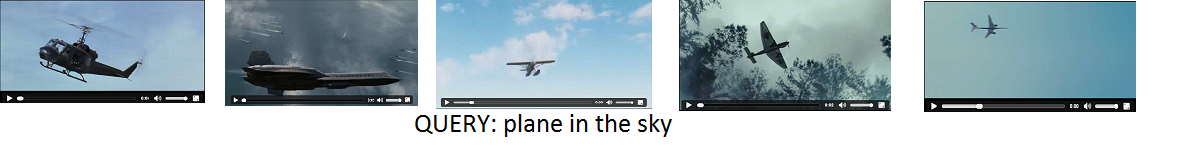} \\
\includegraphics[width=0.95\textwidth]{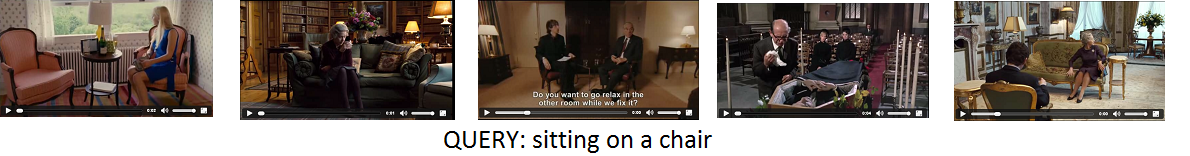} \\
\includegraphics[width=0.95\textwidth]{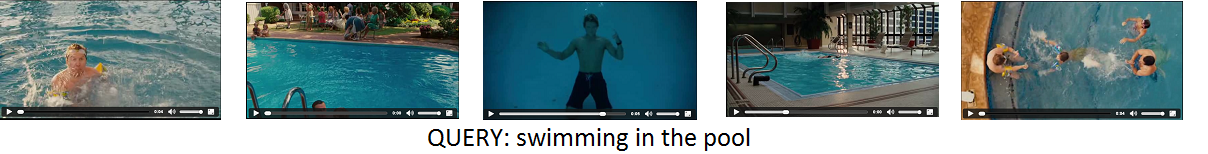} \\
\end{center}
\vspace{-0.1in}
\label{fig:long3}
 \caption{Top 5 (from left to right) phrase-based video search results with natural language  using model \textbf{M2} (\textit{C+L'16}). }
\vspace{-0.2in}
\end{figure*}

\section{Model details}\label{App:AppendixB}
Detail about sentence and video encoding and ranking loss that we used in our models.

\subsection{Video and sentence encoding}
\label{input}
 Each sentence is a sequence of words. The length of word sequences are variable from one sentence to the next. Each word is initially encoded using GloVe \cite{pennington2014glove} distributed word vector representation that has been pre-trained on a 6 billion word corpus that includes Wikipedia and Gigaword 5th edition datasets. Each sentence is therefore represented by a matrix:
\begin{equation}
\label{equ:input}
\textbf{S} = \left[\mathbf{w}_{1}, ..., \mathbf{w}_{N} \right] \in \mathbb{R}^{N \times d_{\mathbf{w}}} \end{equation}
of $N$ word feature vectors, and each vector has $d_{w} = 300$ dimension in our GloVe representation. 
In model $M1$, fixed length representation of the sentence can then be obtained as simple average of the word representations, $\frac{1}{N} \sum_{i=1}^{N} \mathbf{w}_{i}$. The sentence representation is then fed into a linear transformation and the absolute values of transformation output is taken as semantic representation of the full sentence. In model $M2$ and $M3$, the sentence representation $\mathbf{S}$ is fed into an LSTM which results in a sequence of hidden states $\left[\mathbf{h}_{1}, ..., \mathbf{h}_{N} \right] \in \mathbb{R}^{N \times d_{h}}$, where $d_h$ is the dimension of the LSTM hidden state. The absolute values of the output of the last hidden state $\mathbf{h}_{N}$ is taken as semantic representation of the full sentence.

A video is represented as sequence of frames which are sampled every $10$ video frames (images are treated as single frame videos). We extract frame features using pre-trained VGG-19 \cite{Simonyan14c} convolutional neural network (CNN) in Caffe \cite{jia2014caffe}. VGG-19 network \cite{Simonyan14c} is trained for object classification on ImageNet, to classify among $1,000$ object categories. Video sequence length varies from one sequence to the next, which results in a matrix  representation:
\begin{equation}
\label{equ:input}
\textbf{V} = \left[\mathbf{v}_{1}, ..., \mathbf{v}_{M} \right] \in \mathbb{R}^{M \times d_{\mathbf{v}}} \end{equation}
of $M$ video frame feature vectors, each of dimensionality $ d_{\mathbf{v}}$. VGG-19, proposed in \cite{Simonyan14c}, has 16 convolutinal layers and 3 fully connected layers, followed by a softmax output layer. For our experiments we extract second Fully-Connected layer (FC7), therefore $d_{\mathbf{v}} = 4096$. Fixed length representation of the video can then be obtained as simple average of the frame-based representations, $\frac{1}{M} \sum_{i=1}^{M} \mathbf{v}_{i}$, a weighted average obtained using soft attention or using an LSTM encoding.


\subsection{Learning to rank}
\label{LR}

{\bf Two-level partial order similarity:} 
We use the negative order-violation penalty, proposed in \cite{IvanVendrov2015}, for similarity metric, which is defined as    
\begin{equation}
\label{eq:sim}
S(c,v) = - \left\| max\left(0,c-v\right)\right\|^{2} 
\end{equation}
%
using order violating penatly, where $c$ is either an embedding ( for absolute values of simple average: $| \mathbf{W_{word}} \cdot \frac{1}{M} \sum_{i=1}^M \mathbf{v}_i | $, with embedding matrix $\mathbf{W_{word}}$)) or  
$c = |\mathbf{h}_N|$ denotes the absolute values of last hidden state of the language LSTM and $v$ is an embedding (for absolute values of simple/weighted average: $|\mathbf{W_{video}} \cdot \frac{1}{M} \sum_{i=1}^M \mathbf{v}_i | $, with embedding matrix $\mathbf{W_{video}}$). $\mathbf{W_{word}}$, $\mathbf{W_{video}}$, and $\mathbf{h}_N$ are netowrk parameters that are learned during training phase. 

The advantage of this similarity distance is that it is asymmetric, compared to cosine similarity which is symmetric, so it can capture the relatedness of captions with very different lengths that describe the same visual content but in different levels of detail (detailed discussion is given in \cite{IvanVendrov2015}). 

\noindent
{\bf Pairwise ranking loss:}
As baseline, we use the standard loss that has been used in multiple works for image retrieval task in \cite{kiros15tacl,IvanVendrov2015} and is defined as
\begin{equation}
\label{ranking}
{\sum_{(c,v)}\bigg( \sum_{c^{'}} max\left\{0,\alpha-S(c,v)+S(c^{'},v)\right\} + \sum_{v^{'}} max\left\{0,\alpha-S(c,v)+S(c,v^{'})\right\} \bigg)  }\\
\end{equation}
where $(c, v)$ is an encoding for the ground-truth video/sentence pair, $c^{'}$ (contrastive captions) are captions that do not belong to $v$ and $v^{'}$ (contrastive videos) are video/images that are not captioned by $c$; $\alpha$ denotes a margin hyperparameter and $S$ is the similarity function in Eq.~\ref{eq:sim}.

\noindent
{\bf Annotation ranking loss:}
The first term in the above pairwise ranking \ref{ranking}, is a hinge loss that promote a pair of  video and ground-truth caption is required to score higher than all other pairings of same video with contrastive captions by a margin of $\alpha$. We name this annotation ranking loss and we show for multi-choice test, training models with this loss has better performance compared to pairwise ranking loss. Annotation ranking loss is defined as  
\begin{equation}
\label{ranking2}
 {\sum_{(c,v)}\bigg( \sum_{c^{'}} max\left\{0,\alpha-S(c,v)+S(c^{'},v)\right\}\bigg) }\\ 
\end{equation}
\end{document}